\documentclass{article}

\PassOptionsToPackage{numbers, compress}{natbib}

\usepackage[preprint]{neurips_2023}

\usepackage[utf8]{inputenc} 
\usepackage[T1]{fontenc}    
\usepackage{hyperref}       
\usepackage{url}            
\usepackage{booktabs}       
\usepackage{amsfonts}       
\usepackage{nicefrac}       
\usepackage{microtype}      
\usepackage{xcolor}         
\usepackage{graphicx}
\usepackage{float}
\usepackage{placeins}
\usepackage{multirow}
\usepackage{caption}

\title{Open Data on GitHub: Unlocking the Potential of AI}

\author{%
  Anthony Cintron Roman \\
  Microsoft AI for Good Research Lab\\
  \texttt{anthony.cintron@microsoft.com} \\
  \And
  Kevin Xu \\ 
  GitHub
  \And
  Arfon Smith \\ 
  GitHub
  \And
  Jehu Torres \\ 
  Microsoft AI for Good Research Lab
  \And
  Caleb Robinson \\ 
  Microsoft AI for Good Research Lab
  \And
  Juan M. Lavista Ferres \\ 
  Microsoft AI for Good Research Lab
}

\begin{document}

\maketitle

\begin{abstract}
GitHub is the world’s largest platform for collaborative software development, with over 100 million users. GitHub is also used extensively for open data collaboration, hosting more than 800 million open data files, totaling 142 terabytes of data. This study highlights the potential of open data on GitHub and demonstrates how it can accelerate AI research. We analyze the existing landscape of open data on GitHub and the patterns of how users share datasets. Our findings show that GitHub is one of the largest hosts of open data in the world and has experienced an accelerated growth of open data assets over the past four years. By examining the open data landscape on GitHub, we aim to empower users and organizations to leverage existing open datasets and improve their discoverability -- ultimately contributing to the ongoing AI revolution to help address complex societal issues. We release the three datasets that we have collected to support this analysis as open datasets at \url{https://github.com/github/open-data-on-github}.
\end{abstract}

\section{Introduction}
Artificial intelligence (AI) is transforming the world by driving digital innovation, fostering increased experimentation, enhancing efficiency, and accelerating progress across various sectors~\cite{aiindexreport}. Open data plays a crucial role in this transformation, as it offers access to large amounts of information that is essential for developing AI models. This has led to significant advancements in areas such as healthcare, science, education, and environmental conservation~\cite{norouzzadeh2018automatically,jumper2021highly,he2019practical}.

GitHub is one of the preferred choices for developers to share and collaborate on code globally, with over 100 million users and a 27\% year-over-year growth rate~\cite{GitHub_hundred_million,GitHub_Octoverse_2022}. By our analysis, it also ranks among the largest open data platforms in the world, with more than 11 million repositories containing over 800 million open data files.

In 2020, Microsoft launched the Open Data Campaign to assist organizations in harnessing the power of open data and closing the data divide~\cite{Microsoft_Open_Data_Campaign}. To further this objective, we have undertaken this study which analyzes GitHub public repository metadata to provide a deeper understanding of the open data landscape on the GitHub platform. Specifically, using internal access to the data backend for GitHub public APIs~\cite{GitHub_Rest_Search}, we collect a dataset containing the counts of different file types on GitHub that is broken down by username/organization and year the file was added to GitHub. We also analyze the various file types by the license of the repository to which they were added and, finally, a list of websites that host open data. This can provide valuable insights to users and organizations on GitHub, enabling them to maximize the potential of these datasets and enhance their discoverability.

To summarize, we make the following contributions:
\begin{itemize}
    \item We create three open datasets of metadata from over 11 million GitHub repositories that contain structured data files (Section \ref{sec:data_collection_methods})
    \item We provide insights about the open data landscape of the GitHub platform (Section \ref{sec:findings_and_observations})
    \item We discuss our recommendations to maximize the potential of open data on GitHub for accelerating AI research (Section \ref{sec:discussion}) 
\end{itemize}

\section{Data Collection} \label{sec:data_collection_methods}
We curate three open datasets: the \textbf{data counts} dataset -- a tabulation of different data file types on GitHub broken down by user name/organization name and year added; the \textbf{license counts} dataset -- a count of data file types by license and year added; and the \textbf{open data websites} dataset -- a list of websites that host open data with a count of the number of datasets that they host. We have released each dataset licensed under CDLA-Permissive-2.0, available here \url{https://github.com/github/open-data-on-github}. The following subsections describe the methods used to collect each dataset and their associated limitations.

\subsection{Data counts datasets} \label{subsec:github_repo_metadata}
Our \textbf{data counts} dataset consists of a count of the different types of data files on GitHub broken down by the user or organization that posted the file, as well as the year that the file was added to GitHub. The username column has been anonymized for privacy reasons -- a total of 9.8 million rows.  This allows us, for example, to see the users or organizations that have posted the largest numbers of a certain type of file, and determine the rate at which these files are being posted over time.

We collected this dataset programmatically using GitHub's internal data services, which utilize the same backend data as the GitHub public APIs and GitHub search. As not all files on GitHub can be considered data files, we conducted a manual random sampling process of the repositories to assess the usage of each file extension in our subset and to determine the specific use of extensions like JSON. This allowed us to create an exclusion list and establish more precise criteria to minimize the occurrence of false positives. Therefore, during the tabulation process, we filtered based on the following criteria:
\begin{description}
    \item[Included Extensions] We included only the following extensions: json, tsv, tfrecords, gvf, parquet, avro, geojson, pdb, csv, tif, xlsx, fasta, vcf, fastq, gvcf, shp, shx, dbf, kml, wav, dcm, nii, tiff, xls, sbml, biopax2, sbgn.
    \item[Excluded File Names] We excluded the following file names: spec.json, angular.json, package-lock.json, package.json, tsconfig.app.json, tsconfig.json, content.json, .bower.json, .cargo-checksum.json, .eslin-trc.json, app.json, appsettings.Development.json, appsettings.json, appsettings.json, asset-manifest.json, azure-deploy.json, azuredeploy.parameters.json, block.json, bower.json, build.json, cache.json, component.json, composer.json, contents.json, details.json, index.json, manifest.json, messages.json, meta.json, metadata.json, mix-manifest.json, module.json, prereq.azuredeploy.json, prereq.azuredeploy.parameters.json, project.assets.json, properties.json, settings.json, tsconfig.spec.json, tslint.json, cmakelists.json, format.json, authz.json, passwd.json, svnserve.conf.json
    \item[Excluded File paths] We excluded the following file paths: *node\_modules*, *vendor*, ``.'', *config*, *conf*, *bin*, *conf*, *modules*, *app*, *framework*, *src*.
    \item[Excluded Users] After using the file-based filters described above, we also excluded users that met both of the following conditions: (1) the user only created .json files; and (2) the user created fewer than 1000 .json files.
\end{description}

All the GitHub data in this study is based on a snapshot of public repository metadata from 2008 through  March 2023.

\paragraph{Limitations of analysis} This corpus represents the open data published on GitHub based on specific parameters. For instance, our study only includes data files with a certain set of file extensions, and only public, non-forked repositories are included. Additionally, we filtered out certain file paths regardless of content, and have omitted compressed versions such as csv.gz. Moreover, a large portion of the data files are JSON files, which are used for a variety of purposes (e.g. as application configurations). We attempt to control for this limitation with the filtering step, but understand that we still may not be able to cover every scenario. Lastly, due to the sheer number of data files and lack of metadata about the data files, we are unable to validate the content of each one of them for accuracy.

Despite these challenges, we believe our numbers are conservative, as we have filter and excluded paths and file extensions such as ``txt'', and others, due to the difficulty of distinguishing between data and non-data files. As a result, organizations like \url{https://github.com/GITenberg}, which have published over 43k ebooks in the txt format, that could be beneficial in natural language processing and other machine learning applications, have been omitted from our analysis.

\subsection{License Counts Dataset}
Our license count dataset contains a count of files by file format, license, and year added to GitHub. The licenses are assigned at the GitHub repository level. This dataset includes the same filters described in Section~\ref{subsec:github_repo_metadata}.

\paragraph{Limitations of analysis}
We assume that the license assigned to the repository accurately reflects the content that is being shared and distributed.

\subsection{Open Data Websites}
To extend our analysis, we manually curated a dataset of 205 open data websites that includes a website's classification (either governmental, non-profit, or private) and the number of datasets hosted on the website. To identify ``government'' websites hosting open data, we used a list hosted by data.gov~\cite{Open_Government_List} as well as a manual web search. To identify ``private'' and ``non-profit'' open data websites we conducted a manual web search, used Google data search, and through references in ``Google Dataset Search By the Numbers''~\cite{benjelloun2020google}. This corpus is up to date as of February 11, 2023 and, to our knowledge, the largest collection of websites that host open data to date.

\paragraph{Limitations of analysis}
Our open data websites dataset is limited to websites that directly host open data, are accessible through web search, are hosted in English, and are hosted as of February 11, 2023. Additionally, we acknowledge that the manual search of some of the websites included in this dataset is subject to the biases of the search engines used. Therefore, the limitations include:
\begin{itemize}
\item Websites that directly host open data and do not include other sources of open data such as APIs, databases, dashboards or metadata aggregators such as Google data search.
\item Open data Websites that are accessible through web search and do not include websites that are not indexed by search engines.
\item Open data Websites that are hosted in English.
\item Open data Websites that existed as of February 11, 2023 and do not include websites that will be hosted in the future.
\end{itemize}

\section{Empirical Findings and Observations} 
\label{sec:findings_and_observations}
In this section we discuss the open data landscape of the GitHub platform (Section 3.1), how data is published, shared and licensed (Section 3.2) and finally the discoverability of datasets on GitHub (Section 3.3). 
\begin{figure}[t]
    \centering
    \includegraphics[width=0.8\linewidth]{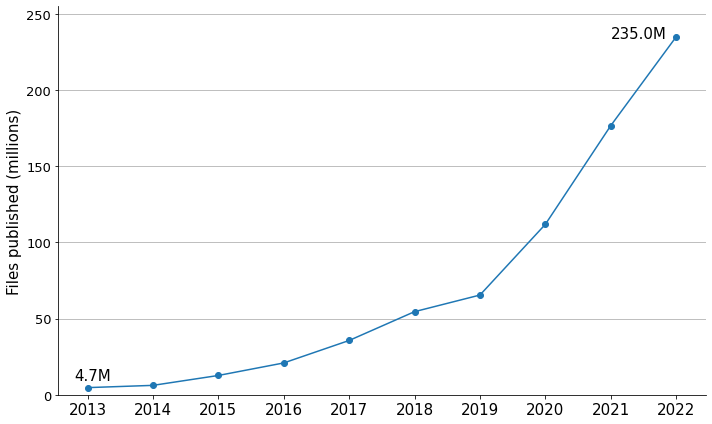}
    \caption{Number of data assets by year added to GitHub from 2013 to 2022.}
    \label{fig:open_data_growth}
\end{figure}
\subsection{Open Data on GitHub}
Based on our data counts dataset, which spans data files added to GitHub from March 26, 2008 to April 1, 2023, GitHub hosts over 11.3 million public repositories containing over 802 million structured and machine-readable data files. Over 4.2 million users publish data files on the platform.

Over the last 10 years (2013-2022) over 723 million data files have been published to the GitHub platform (Fig \ref{fig:open_data_growth}). 57\% of those data files were published in the last 2 years and 81\% in the last 4 years.
Of the 11.3 million repositories containing public data assets, 1.3 million are owned by organizations, while 10 million are owned by individual users. Despite having fewer repositories, organizations account for 31\% of the 802 million data files on GitHub, with 252 million files.

\begin{table}[ht]
    \centering
    \caption{Top 10 data file formats on GitHub}
    \label{tab:dataset_formats}
    \begin{tabular}{@{}lcc@{}}
    \toprule
    \multicolumn{1}{c}{\textbf{File Format}} & \textbf{Data Assets} \\ \midrule
    \ json & 483.7M \\
    \ csv & 168.2M \\
    \ geojson & 44.2M \\
    \ wav & 43.7M \\
    \ pdb & 14.7M \\
    \ tif & 14M \\
    \ xlsx & 7.8M \\
    \ tsv & 5.5M \\ 
    \ fasta & 4.2M \\
    \ parquet & 3.2M \\ \bottomrule
    \end{tabular}
    \end{table}

Of the 802 million data files we considered, 81\% (652 million) are in the JSON and CSV file formats (Table \ref{tab:dataset_formats}), which is unsurprising given their widespread use. However, since JSON files are used for many purposes, including configuration files, we excluded certain file names, file paths, and users, as discussed in Section \ref{subsec:github_repo_metadata}, to filter out non-data files as much as possible. Further, we manually evaluated the files from 45 users that account for 26\% (127 million) of JSON files published and found that 63\% of those files (80 million) were data files, related to, among other things, the transportation, geospatial, and medical domains.

\begin{table}[h]
    \centering
    \caption{Number of data assets on GitHub by research category}
    \label{tab:dataset_list_category}
    \begin{tabular}{@{}lcc@{}}
    \toprule
    \multicolumn{1}{c}{\textbf{Category}} & \textbf{Data Assets} & \textbf{Formats} \\ \midrule
    \ General & 689.3M & json,csv,tsv,xls,xlsx,tiff,tif,parquet,avro,dbf \\
    \ Geospatial & 47.5M & shp,geojson,shx \\
    \ Acoustic & 43.7M & wav \\
    \ Chemistry & 14.7M & pdb \\
    \ Bioinformatics & 4.4M & fasta,fastq,biom,sbgn,sbml \\
    \ MedicalImaging & 2.6M & nii,dcm \\
    \ Genomics & 215K & gvf,gvcf,vcf \\ \bottomrule
    \end{tabular}
    \end{table} 

GitHub hosts a wide range of data assets that can be used for various purposes. Breaking down file format by topic area we observe that, following the general-purpose file formats, geospatial data is the next largest area, with over 47.5 million data files that have GeoJson, SHX, SHP and KML file formats (Table \ref{tab:dataset_list_category}).

\begin{table}[ht]
    \centering
    \caption{Number of data assets for the top ten open data sites}
    \label{tab:dataset_list}
    \begin{tabular}{@{}lcc@{}}
    \toprule
    \multicolumn{1}{c}{\textbf{Website}} & \textbf{Data Assets} & \textbf{Sector} \\ \midrule
    \url{github.com} & 802M Files / 11.3M Repos & private \\
    \url{figshare.com/search} & 1.6M & private \\
    \url{datadiscoverystudio.org/geoportal} & 1.6M & non-profit \\
    \url{data.europa.eu/catalogue-statistics} & 1.3M & government \\
    \url{data.gov.in} & 499K & government \\
    \url{maps.amsterdam.nl/open_geodata} & 372k & government \\
    \url{catalog.data.gov/dataset} & 245K & government \\
    \url{researchdata.ands.org.au} & 203K & government \\
    \url{kaggle.com} & 197K & private \\
    \url{zenodo.org} & 178K & non-profit \\ \bottomrule
    \end{tabular}
    \end{table}

By comparing the GitHub open data landscape to the industry open data landscape, it is clear that GitHub is one of the largest open data platforms in the world (Table \ref{tab:dataset_list}). GitHub offers an impressive variety and volume of data assets. Governments are the second largest open data providers, with 3.9 million data assets available worldwide. Private organizations offer 2 million data assets, while nonprofits offer 1.9 million.

\subsection{Data Publishing}
Data publishing on GitHub can be accomplished in various ways, such as uploading the dataset to the repository, providing links to an external storage solution in a readme file, or a combination of both. When datasets are large, it is often more practical to publish by reference. Additionally, with Git Large File System (LFS), files can be published with a size larger than the default maximum of 100MB. Depending on the type of GitHub account, the maximum file size can be up to 5GB~\cite{GitHub_git_LFS}.

\begin{table}[h]
    \centering
    \caption{Four GitHub accounts that have published over 40 million data files.}
    \label{tab:top4}
    \resizebox{\columnwidth}{!}{%
    \begin{tabular}{@{}lrrrrrrrr@{}}
    \toprule
    \multicolumn{1}{c}{\multirow{2}{*}{\textbf{Owner}}} & \multicolumn{7}{c}{\textbf{File Types}} & \multicolumn{1}{c}{\multirow{2}{*}{\textbf{Total}}} \\ \cmidrule(lr){2-8}
    \multicolumn{1}{c}{} & \multicolumn{1}{c}{\textbf{geojson}} & \multicolumn{1}{c}{\textbf{json}} & \multicolumn{1}{c}{\textbf{tsv}} & \multicolumn{1}{c}{\textbf{wav}} & \multicolumn{1}{c}{\textbf{csv}} & \multicolumn{1}{c}{\textbf{shp}} & \multicolumn{1}{c}{\textbf{tiff}} & \multicolumn{1}{c}{} \\ \midrule
    whosonfirst-data & 24,056,588 & 7,321 & 0 & 0 & 52 & 9 & 0 & 24,063,970 \\
    sfomuseum-data & 8,522,469 & 1,595 & 0 & 0 & 1 & 0 & 0 & 8,524,065 \\
    woeplanet-data & 5,762,402 & 1,239,622 & 444 & 0 & 6 & 0 & 0 & 7,002,474 \\
    FitzwilliamMuseum & 2 & 1,430,103 & 0 & 304 & 35 & 0 & 2 & 1,430,446 \\ \midrule
    \textbf{Total} & 38,341,461 & 2,678,641 & 444 & 304 & 94 & 9 & 2 & \textbf{41,020,955} \\ \bottomrule
    \end{tabular}%
    }
    \end{table}

Although there is a default file size limit of 100MB and overall repository size limit of 5GB when publishing to GitHub repositories~\cite{GitHub_About_Files}, data publishers have been able to share large amounts of data on the platform. For example, FitzwilliamMuseum, sfomuseum-data, woeplanet-data, whosofirst-data and other users and organizations have split their datasets into hundreds and thousands of GitHub repositories, enabling them to publish their data within GitHub’s platform limits, albeit with careful efforts to optimize the git history in some cases~\cite{Whonfirst_Data_Changes}. These four organizations alone have contributed more than 41 million data files (Table \ref{tab:top4}).

\subsection{Data Licensing}
Licensing is also an important part of data publishing and sharing. Open data is based on the idea that anyone can access, use, and share it freely, and the license associated with a dataset can provide important information about what uses its authors intend to enable, what uses are prohibited, and how to provide attribution when the dataset is used in other projects. GitHub attempts to classify repositories based on license if a single, known license type is detected in a well known location, such as a LICENSE file in the root directory of a repository.

\begin{figure}[ht]
    \centering
    \begin{minipage}[t]{0.50\linewidth}
    \centering
    \captionof{table}{Percentage of data assets by license category}
    \label{tab:dataset_permissive_licenses}
    \begin{tabular}{@{}lcc@{}}
    \toprule
    \multicolumn{1}{c}{\textbf{License Category}} & \textbf{Data Assets} & \textbf{Percentage} \\ \midrule
    \ {\color[HTML]{2ca02c} "Open Data" License} & 9.0M & 1.1\% \\
    \ {\color[HTML]{8c564b} Unidentified License} & 59.9M & 7.5\% \\
    \ {\color[HTML]{ff7f0e} Other Open Licenses} & 124.9M & 15.6\% \\
    \ {\color[HTML]{1F77B4} No License Detected} & 609.0M & 75.9\% \\
    \bottomrule
    \end{tabular}
    \end{minipage}
    \begin{minipage}[t]{0.45\linewidth}
    \centering
    \strut\vspace*{-\baselineskip}\newline\includegraphics[width=0.75\linewidth]{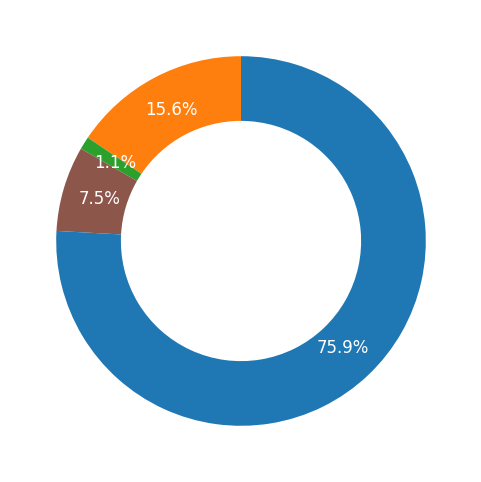}
    \end{minipage}
    \end{figure}

Out of the 802 million data files on GitHub, over 9 million have a conformant “open data” license as defined by the Open Knowledge Foundation (OKF)~\cite{OKF_Conformant_License_Def} (one of CC0-1.0, PDDL-1.0, CC-BY-4.0, CC-BY-SA-4.0, ODbL-1.0, or O-UDA-1.0 -- note that ODC-By-1.0 was omitted because GitHub’s automated system does not currently identify ODC-By-1.0), and over 124 million have a different open license detected and identified by GitHub’s automated systems. However, over 609 million data files do not have a license, and over 59 million have a license that GitHub’s automated systems could detect, but could not identify, either because there were multiple licenses associated with the repository, or the text of the detected license was not sufficiently similar to a known license file (Table \ref{tab:dataset_permissive_licenses}). For the purpose of this study, these files are still considered public data assets without restrictions of use. Nevertheless, it is essential to assign a license to repositories in order to give users assurance that the dataset can be utilized ~\cite{GitHub_Adding_Licenses,ODI_Licenses}.

\subsection{Data Discovery}
At present, GitHub provides features such as Collections, Topics, and a standard search function to locate assets within the platform. These features are primarily designed for code search but can also be utilized to find repositories containing data files. GitHub Topics enable users to explore repositories related to specific subject areas. However, since topics are free-form, users can tag their repositories with any topic they choose, resulting in a lack of standardization for open data topics~\cite{GitHub_Topics}.

On the other hand, GitHub Collections allow users to discover hand-picked repositories that share a common theme, with the key difference being that collections are curated~\cite{GitHub_Collections}. Despite these features, only 0.001\% of repositories and 0.04\% of data assets published on GitHub can be found through Collections, Topics, and the standard code search functions. Currently, there is one Open Data Collection consisting of 8 repositories with approximately 24.3 million data assets, and 7 data Topics encompassing 14,564 repositories with around 5.5 million data assets.

As a result, it is crucial to identify and tag data repositories using these features to enhance their discoverability and make it easier for users to find relevant information.

\section{Discussion} \label{sec:discussion}

\subsection{Accelerating AI Research}
In March 2023, Google released a dataset on GitHub as part of a paper that demonstrated the re-training of diffusion text to image models to include ``personalization'' on the generated images~\cite{ruiz2022dreambooth}. This dataset is a great asset for AI practitioners working on text to image research and a great example of datasets available on GitHub for the advancement of AI.

In addition to the dataset released by Google, there are numerous other datasets available on GitHub that have significantly contributed to the advancement of AI research. These datasets cover a wide range of topics and domains, providing AI practitioners with valuable resources to develop and improve their models. For instance:

\begin{description}
    \item[SQuAD (Stanford Question Answering Dataset)] SQuAD is a popular dataset for natural language processing and machine comprehension tasks. It consists of over 100,000 question-answer pairs based on Wikipedia articles~\cite{rajpurkar2016squad}. The dataset is available on GitHub at \url{https://github.com/rajpurkar/SQuAD-explorer}.
    \item[OpenAI GPT-2 Dataset] OpenAI released the GPT-2 dataset, which is a large-scale language model trained on diverse internet text. This dataset has been influential in the development of natural language processing and generation models~\cite{radford2019language}. The dataset is available on GitHub, and more information can be found at \url{https://github.com/openai/gpt-2-output-dataset}.
    \item[COCO (Common Objects in Context)] COCO is a widely used dataset for object detection, segmentation, and captioning tasks. It contains over 200,000 labeled images with 1.5 million object instances across 80 categories~\cite{lin2014microsoft}. The COCO website is hosted on GitHub (\url{https://github.com/cocodataset/cocodataset.github.io}), an example of sharing data by reference on GitHub.
\end{description}

These examples demonstrate the importance of open data, and the role GitHub plays in hosting and sharing datasets for AI research. Open data is of tremendous value for AI practitioners, is the ability to produce better results and produce new innovations. Datasets hosted on GitHub can enable AI researchers to access larger quantities of data than currently available, enabling improved AI models. This provides a powerful tool for advancing research in the AI field.

\subsection{Future Work}

In this work, we explore the landscape of open data on GitHub using several filters based on file metadata such as extensions, names, paths, and counts per user. As discussed above in their respective limitations sections, each of these filters are imperfect given the wide range of purposes for which users push files to GitHub or choose a particular file extension/name/path. Future work should explore using methods to classify repositories as containing open data files based on repository content data, such as the text of the documentation files within the repository. Such content-based classification approaches would enable analysis of truly general-purpose extension types like .txt, which, as mentioned previously, were excluded entirely from this work.

\section{Conclusion}

GitHub is the home to over 802 million data files posted by over 4 million users and organizations across a variety of domains. The rate that new data files are being posted on GitHub is increasing -- 57\% of the data files have been posted in the past two years.

This large amount of data on GitHub has the potential to significantly advance AI research by providing researchers and practitioners with access to a vast and growing amount of data across different domains. The platform itself can be used to foster collaboration and promote the development of more sophisticated AI models, as evidenced by community AI initiatives hosted on GitHub like LAION.AI (e.g., ~\url{https://github.com/LAION-AI/Open-Assistant} and ~\url{https://github.com/LAION-AI/CLIP-based-NSFW-Detector}. Other studies have highlighted the importance of open data in AI research, emphasizing its role in driving innovation and improving productivity~\cite{mckinsey_opendata,healthcare9070827,BRINKHAUS2023102542}. 

Access to large datasets is crucial for the scaling of AI models, and by hosting open data on GitHub, researchers can easily access and utilize these datasets, ultimately leading to advancements in the field. Furthermore, GitHub's collaborative environment promotes an open and transparent AI research community. Data and collaboration are essential for driving innovation in AI research, as they allow researchers to build upon each other's work and develop new ideas more efficiently.

Finally, open data on GitHub can also contribute to the democratization of AI research, as it allows researchers from various backgrounds and institutions to access and utilize the same datasets. This can help level the playing field and promote more diverse perspectives in AI research, ultimately leading to more innovative and inclusive AI solutions.

\bibliographystyle{plain}
\bibliography{citations}

\begin{thebibliography}{10}

\bibitem{benjelloun2020google}
Omar Benjelloun, Shiyu Chen, and Natasha Noy.
\newblock Google dataset search by the numbers.
\newblock In {\em The Semantic Web--ISWC 2020: 19th International Semantic Web
  Conference, Athens, Greece, November 2--6, 2020, Proceedings, Part II}, pages
  667--682. Springer, 2020.

\bibitem{BRINKHAUS2023102542}
Henning~Otto Brinkhaus, Kohulan Rajan, Jonas Schaub, Achim Zielesny, and
  Christoph Steinbeck.
\newblock Open data and algorithms for open science in ai-driven molecular
  informatics.
\newblock {\em Current Opinion in Structural Biology}, 79:102542, 2023.

\bibitem{ODI_Licenses}
Leigh Dodds.
\newblock {Publisher’s Guide to Open Data Licensing}.
\newblock
  \url{https://theodi.org/article/publishers-guide-to-open-data-licensing/},
  December 2013.

\bibitem{GitHub_hundred_million}
Thomas Dohmke.
\newblock {100 million developers and counting}.
\newblock
  \url{https://github.blog/2023-01-25-100-million-developers-and-counting},
  2023.

\bibitem{OKF_Conformant_License_Def}
Open~Knowledge Foundation.
\newblock {Conformant Licenses - Open Definition - Defining Open in Open Data,
  Open Content and Open Knowledge}.
\newblock \url{https://opendefinition.org/licenses}, 2023.

\bibitem{GitHub_Collections}
GitHub.
\newblock {GitHub Collections}.
\newblock \url{https://github.com/collections}.

\bibitem{GitHub_Octoverse_2022}
GitHub.
\newblock {Octoverse 2022}.
\newblock \url{https://octoverse.github.com}, 2022.

\bibitem{GitHub_git_LFS}
GitHub.
\newblock {About Git Large File Storage}.
\newblock
  \url{https://docs.github.com/en/repositories/working-with-files/managing-large-files/about-git-large-file-storage},
  2023.

\bibitem{GitHub_About_Files}
GitHub.
\newblock {About Large Files On GitHub}.
\newblock
  \url{https://docs.github.com/en/repositories/working-with-files/managing-large-files/about-large-files-on-github#repository-size-limits},
  2023.

\bibitem{GitHub_Adding_Licenses}
GitHub.
\newblock {Adding a license to a repository}.
\newblock
  \url{https://docs.github.com/en/communities/setting-up-your-project-for-healthy-contributions/adding-a-license-to-a-repository},
  2023.

\bibitem{GitHub_Topics}
GitHub.
\newblock {Classifying your repository with topics}.
\newblock
  \url{https://docs.github.com/en/repositories/managing-your-repositorys-settings-and-features/customizing-your-repository/classifying-your-repository-with-topics},
  2023.

\bibitem{GitHub_Rest_Search}
GitHub.
\newblock {Search}.
\newblock \url{https://docs.github.com/en/rest/search}, 2023.

\bibitem{Open_Government_List}
U.S Government.
\newblock {Open Government}.
\newblock \url{https://data.gov/open-gov}.

\bibitem{he2019practical}
Jianxing He, Sally~L Baxter, Jie Xu, Jiming Xu, Xingtao Zhou, and Kang Zhang.
\newblock The practical implementation of artificial intelligence technologies
  in medicine.
\newblock {\em Nature medicine}, 25(1):30--36, 2019.

\bibitem{jumper2021highly}
John Jumper, Richard Evans, Alexander Pritzel, Tim Green, Michael Figurnov,
  Olaf Ronneberger, Kathryn Tunyasuvunakool, Russ Bates, Augustin
  {\v{Z}}{\'\i}dek, Anna Potapenko, et~al.
\newblock Highly accurate protein structure prediction with alphafold.
\newblock {\em Nature}, 596(7873):583--589, 2021.

\bibitem{lin2014microsoft}
Tsung-Yi Lin, Michael Maire, Serge Belongie, James Hays, Pietro Perona, Deva
  Ramanan, Piotr Doll{\'a}r, and C~Lawrence Zitnick.
\newblock Microsoft coco: Common objects in context.
\newblock In {\em Computer Vision--ECCV 2014: 13th European Conference, Zurich,
  Switzerland, September 6-12, 2014, Proceedings, Part V 13}, pages 740--755.
  Springer, 2014.

\bibitem{mckinsey_opendata}
James Manyika, Michael Chui, Peter Groves, Diana Farrell, Steve Van~Kuiken, and
  Elizabeth Almasi~Doshi.
\newblock {Open data: Unlocking innovation and performance with Liquid
  Information}.
\newblock
  \url{https://www.mckinsey.com/capabilities/mckinsey-digital/our-insights/open-data-unlocking-innovation-and-performance-with-liquid-information},
  2013.

\bibitem{aiindexreport}
Nestor Maslej, Loredana Fattorini, Erik Brynjolfsson, John Etchemendy, Katrina
  Ligett, Terah Lyons, James Manyika, Helen Ngo, Juan Carlos~Niebles, Vanessa
  Parli, Yoav Shoham, Russell Wald, Jack Clark, and Raymond Perrault.
\newblock The ai index 2023 annual report.
\newblock {\em AI Index Steering Committee, Institute for Human-Centered AI,
  Stanford University, Stanford, CA}, April 2023.

\bibitem{norouzzadeh2018automatically}
Mohammad~Sadegh Norouzzadeh, Anh Nguyen, Margaret Kosmala, Alexandra Swanson,
  Meredith~S Palmer, Craig Packer, and Jeff Clune.
\newblock Automatically identifying, counting, and describing wild animals in
  camera-trap images with deep learning.
\newblock {\em Proceedings of the National Academy of Sciences},
  115(25):E5716--E5725, 2018.

\bibitem{healthcare9070827}
Tania Pereira, Joana Morgado, Francisco Silva, Michele~M. Pelter, Vasco~Rosa
  Dias, Rita Barros, Cláudia Freitas, Eduardo Negrão, Beatriz Flor~de Lima,
  Miguel Correia~da Silva, António~J. Madureira, Isabel Ramos, Venceslau
  Hespanhol, José~Luis Costa, António Cunha, and Hélder~P. Oliveira.
\newblock Sharing biomedical data: Strengthening ai development in healthcare.
\newblock {\em Healthcare}, 9(7), 2021.

\bibitem{radford2019language}
Alec Radford, Jeffrey Wu, Rewon Child, David Luan, Dario Amodei, Ilya
  Sutskever, et~al.
\newblock Language models are unsupervised multitask learners.
\newblock {\em OpenAI blog}, 1(8):9, 2019.

\bibitem{rajpurkar2016squad}
Pranav Rajpurkar, Jian Zhang, Konstantin Lopyrev, and Percy Liang.
\newblock Squad: 100,000+ questions for machine comprehension of text.
\newblock {\em arXiv preprint arXiv:1606.05250}, 2016.

\bibitem{ruiz2022dreambooth}
Nataniel Ruiz, Yuanzhen Li, Varun Jampani, Yael Pritch, Michael Rubinstein, and
  Kfir Aberman.
\newblock Dreambooth: Fine tuning text-to-image diffusion models for
  subject-driven generation.
\newblock {\em arXiv preprint arXiv:2208.12242}, 2022.

\bibitem{Whonfirst_Data_Changes}
thisisaaronland.
\newblock {Upcoming changes to Who's On First administrative data}.
\newblock \url{https://whosonfirst.org/blog/2019/05/09/changes/}, May 2019.

\bibitem{Microsoft_Open_Data_Campaign}
Jennifer Yokoyama.
\newblock {Open Data Campaign: Exploring the power of open data}.
\newblock
  \url{https://blogs.microsoft.com/on-the-issues/2020/09/22/open-data-covid-19-climate-change},
  2020.

\end{thebibliography}

\end{document}